\documentclass{article}
\usepackage{spconf,amsmath,graphicx}
\usepackage{amsfonts}
\usepackage{booktabs}
\usepackage{multirow}
\usepackage{bm}
\usepackage[linesnumbered, lined, ruled]{algorithm2e}

\title{LEDNET: a lightweight encoder-decoder network for \\ real-time semantic segmentation}
\name{Yu Wang$^{1}$, Quan Zhou$^{1, 2, }$\sthanks{Corresponding author: Quan Zhou, quan.zhou@njupt.edu.cn. This work is partly supported by NSFC (No. 61876093, 61701258, 61701252, 61671253), NSFJS (No. BK20181393, BK20170906), NSF (No. IIS-1302164), and Huawei Innovation Research Program (HIRP2018).}, Jia Liu$^{1}$, Jian Xiong$^{1}$, Guangwei Gao$^{3}$, Xiaofu Wu$^{1}$, and Longin Jan Latecki$^{4}$}
\address{$^{1}$National Engineering Research Center of Communications and Networking, \\ Nanjing University of Posts \& Telecommunications, P.R. China.\\
$^{2}$Key Lab. of Broadband Wireless Communications and Sensor Network Technology, \\ Nanjing University of Posts \& Telecommunications, P.R. China.\\
$^{3}$Institute of Advanced Technology, Nanjing University of Posts \& Telecommunications, P.R. China.\\
$^{4}$Department of Computer and Information Sciences, Temple University, Philadelphia, USA.\\}
\begin{document}
%\ninept
%
\maketitle
\begin{abstract}

The extensive computational burden limits the usage of CNNs in mobile devices for dense estimation tasks. In this paper, we present a lightweight network to address this problem, namely \emph{LEDNet}, which employs an \emph{asymmetric} encoder-decoder architecture for the task of real-time semantic segmentation. More specifically, the encoder adopts a ResNet as backbone network, where two new operations, channel split and shuffle, are utilized in each residual block to greatly reduce computation cost while maintaining higher segmentation accuracy. On the other hand, an attention pyramid network (APN) is employed in the decoder to further lighten the entire network complexity. Our model has less than 1M parameters, and is able to run at over 71 FPS in a single GTX 1080Ti GPU. The comprehensive experiments demonstrate that our approach achieves state-of-the-art results in terms of speed and accuracy trade-off on CityScapes dataset.

\end{abstract}
\section{Introduction}
\label{sec:intro}

Recently, building deeper and larger convolutional neural networks (CNNs) is a primary trend for solving scene understanding tasks \cite{Girshick2014rich,He2016deep,going2015szegedy,long2017fully,Chen2016deeplab}. The most accurate CNNs usually have hundreds of convolutional layers and thousands of feature channels. In spite of achieving higher performance, these advances are at the sacrifice of running time and speed. Especially in the context of many real-world scenarios, such as augmented reality, robotics, and self-driving car to name of few, the computationally cheap networks with smaller size are often required to carry out online estimation in a timely fashion. Therefore, those accurate networks requiring enormous resources are not suitable for computationally limited mobile platforms (e.g., drones, robots, and smartphones), which have limited energy overhead, restrictive memory constraints, and reduced computational capabilities. Such kind of limitation is particularly prominent on the computationally heavy task of semantic segmentation \cite{long2017fully,Chen2016deeplab,Guosheng2017RefineNet,Badrinarayanan2015Segnet,zhao2017pyramid}, where the goal here is to assign a semantic category label for each image pixel.

In order to overcome this problem, many lightweight style networks have been designed to balance the segmentation accuracy and implementing efficiency, which are roughly divided into two categories: network compression \cite{Chen2015compress,Han2016deep,Wu2016quantized,Rastegari2016xnor} and convolution factorization \cite{Paszke2016enet,Romera2018erfnet,Howard2017mobile}. The first category prefers to reduce inference computation by compressing pre-trained networks, including hashing \cite{Chen2015compress}, pruning \cite{Han2016deep}, and quantization \cite{Wu2016quantized,Rastegari2016xnor}. To further remove the redundancy, an alternative approach to lighten CNNs depends on sparse coding theory \cite{wen2016learning,liu2015sparse}. On the contrary, motivated from the convolution factorization principle (CFP) that decomposes a standard convolution into group convolution and depthwise separable convolution \cite{going2015szegedy,Howard2017mobile,Szegedy2016rethinking}, the second category focuses on directly training network with smaller size. For example, ENet \cite{Paszke2016enet} employs ResNet \cite{He2016deep} as backbone to perform efficient inference. Zhao \emph{et al.} \cite{Zhao2018ICnet} propose a cascade network that incorporates high-level label guidance to improve performance. In \cite{Badrinarayanan2015Segnet,Romera2018erfnet,Mehta2018espnet}, a symmetrical encoder-decoder architecture is adopted, which greatly reduce the number of parameters while maintaining the accuracy. Although some work have conducted preliminary research on lightweight architecture networks, pursuing the best accuracy in very limited computational budgets is still an open research question for the task of real-time semantic segmentation.

\begin{figure*}[!t]
\centerline{\includegraphics[width = 0.96\textwidth]{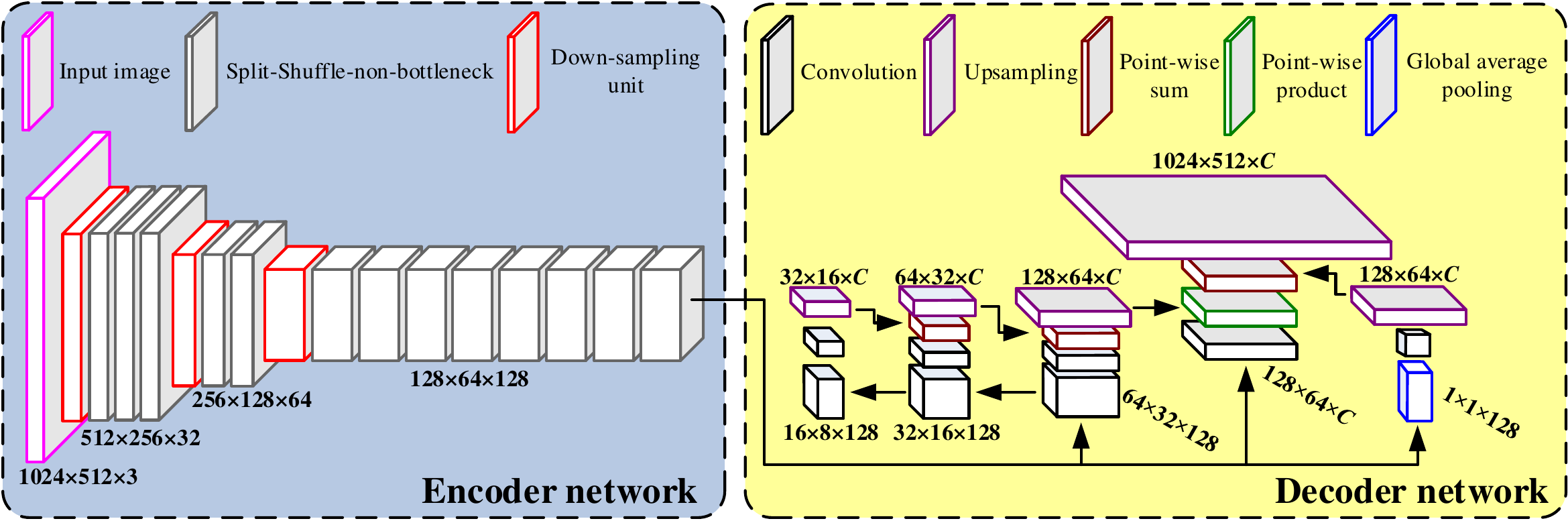}}
\caption{Overall asymmetric architecture of the proposed LEDNet. The encoder employs a FCN-like network, while an APN is adopted in decoder. $C$ denotes the number of classes. Please refer to text for more details. (Best viewed in color)} \label{fig:Overview}
\end{figure*}

In this paper, we aim at solving this trade-off as a whole, without sitting on only one of its sides. We introduce a novel lightweight network called \emph{LEDNet}, adopting an \emph{asymmetric} encoder-decoder architecture for real-time semantic segmentation. As shown in Figure \ref{fig:Overview}, our  LEDNet is composed of two parts: encoder and decoder network. Following CFP, the core unit of encoder is a novel residual module that leverages skip connections and convolutions with channel split and shuffle. While the skip connections allow the convolutions to learn residual functions that facilitate training, the split and shuffle operations enhance the information exchange within the feature channels while maintaining similar computational costs compared to 1D factorized convolutions. In the decoder, instead of complicated dilated convolution \cite{Mehta2018espnet}, we design an attention pyramid network (APN) to extract dense features, where the attention mechanism is utilized to estimate semantic label for each pixel. Our contributions are three-folds: (1) The asymmetrical architecture of our LEDNet leads to the great reduction of network parameters, which accelerates the inference process; (2) The channel split and shuffle operations in our residual layer leverage network size and powerful feature representation. In addition, channel shuffle is also differentiable, which means it can be embedded into network structures for end-to-end training. (3) Attention mechanism of feature pyramid is employed to design APN in our decoder-end, further lightening the complexity of the whole network.

\section{Our Approach}\label{sec:method}

\begin{figure}[!t]
\centerline{\includegraphics[width = 0.5\textwidth]{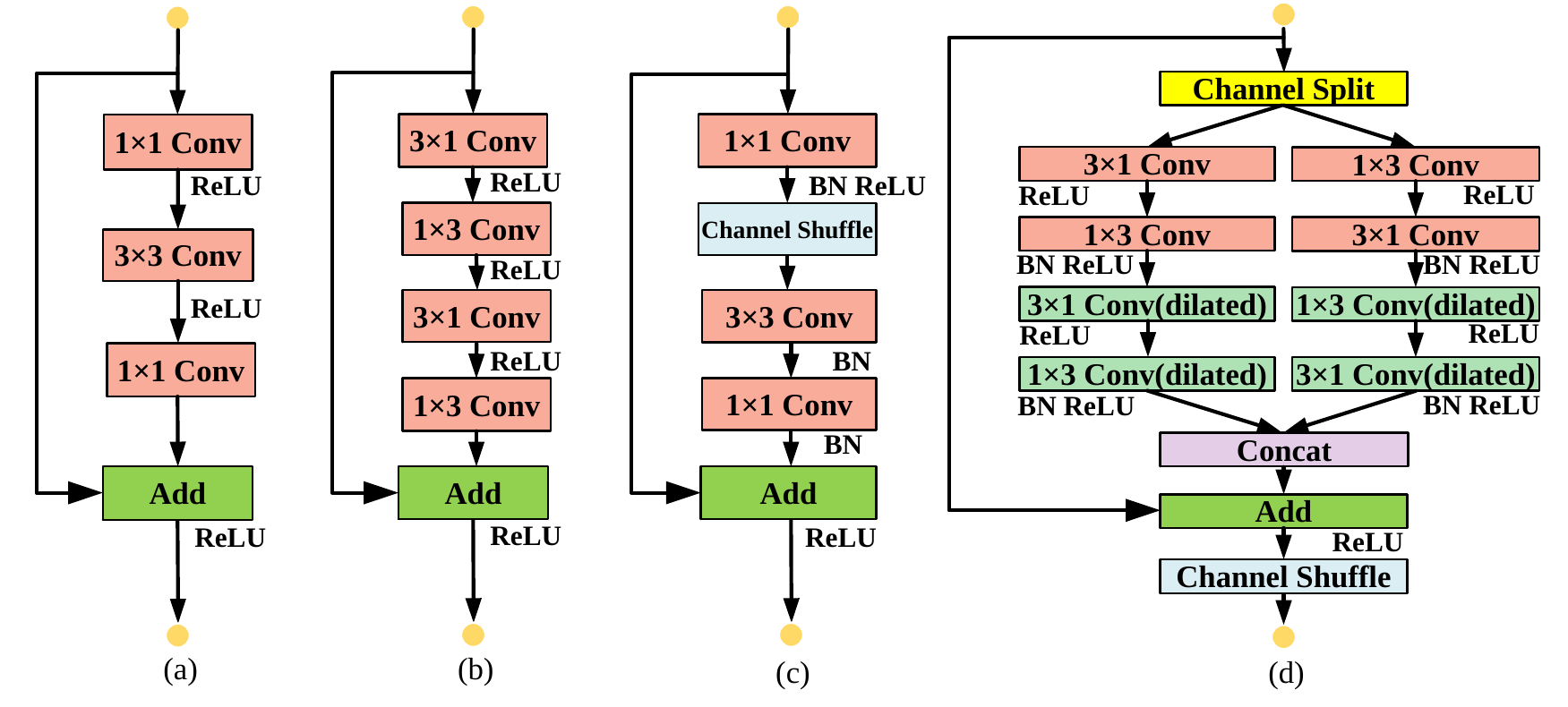}}
\caption{Comparison of different residual layer modules. From left to right are (a) bottleneck \cite{Paszke2016enet,Howard2017mobile}, (b) non-bottleneck-1D \cite{Romera2018erfnet}, (c) ShuffleNet \cite{zhang2018shuffle}, and (d) our SS-nbt module.} \label{fig:ShuffleModule}
\end{figure}

%This section first describes the new residual module, and then elaborates on the detail of our entire LEDNet.

\subsection{Residual Module with Split and Shuffle Operations}

%\noindent \textbf{Residual Module with Split and Shuffle Operations.}
We focus on solving the efficiency limitation that is essentially present in the residual block, which is used in recent accurate CNNs for image classification \cite{He2016deep,xie2017agg,zhang2018shuffle} and semantic segmentation \cite{Guosheng2017RefineNet,Paszke2016enet,Romera2018erfnet}. %To reduce computation, the group convolutions \cite{xie2017agg,zhang2018shuffle} and depthwise convolutions \cite{going2015szegedy,Chen2016deeplab,Howard2017mobile} are adopted as standard steps in residual block. As shown in Figure \ref{fig:ShuffleModule}, the recent work have designed multiple successful instances of this residual layer, including bottleneck, non-bottleneck-1D, and ShuffleNet module.
The recent years have witnessed multiple successful instances of lightweight residual layer \cite{Paszke2016enet,Howard2017mobile},  such as bottleneck (Figure \ref{fig:ShuffleModule} (a)), non-bottleneck-1D (Figure \ref{fig:ShuffleModule} (b)), and ShuffleNet
module (Figure \ref{fig:ShuffleModule} (c)), where the pointwise convolution is widely used.
%For instance, the bottleneck module (Figure \ref{fig:ShuffleModule} (a)) comes from the standard residual layer of ResNet \cite{He2016deep}, which requires less computational resources. Although it is commonly adopted in state-of-the-art networks \cite{Paszke2016enet,Howard2017mobile}, the performance descend drastically when network depth increases. Another two outstanding residual module are non-bottleneck-1D (Figure \ref{fig:ShuffleModule} (b)) and ShuffleNet (Figure \ref{fig:ShuffleModule} (c)), where the first one is a 1D version of bottleneck while the second one utilizes pointwise convolutions (i.e., $1 \times 1$ convolution) in bottleneck structure.
However, the contrary opinion of \cite{zhang2018shuffle} claims that pointwise convolution accounts for most of the computational complexity, which is especially disadvantageous for lightweight models.

To balance performance and efficiency given limited computational budgets, we introduce two simple operators, called channel split and shuffle, in residual layer. We refer to this proposed module as split-shuffle-non-bottleneck(SS-nbt), as depicted in Figure \ref{fig:ShuffleModule} (d). Motivated from \cite{Rastegari2016xnor,Szegedy2016rethinking}, a \emph{split-transform-merge} strategy is employed in the designment of our SS-nbt, approaching the representational power of large and dense layers, but at a considerably lower computational complexity. At the beginning of each SS-nbt, the input is split into two lower-dimensional branches, where each one has half channels of the input. To avoid pointwise convolution, the transformation is performed using a set of specialized 1D filters (e.g., $1 \times3$, $3 \times 1$), and the convolutional outputs of two branches are merged using concatenation so that the number of channels keeps the same. To facilitate training, the stacked output is added with input through the branch of identity mapping. The same channel shuffle operation \cite{zhang2018shuffle} is finally used to enable information communication between two split branches. After the shuffle, the next SS-nbt unit begins. It is clear that our residual module is not only efficient, but also accurate. Firstly, the high efficiency in each SS-nbt allows us to use more feature channels. Secondly, in each SS-nbt unit, the merged feature channels are randomly shuffled, and then join into next unit. This can be regarded as a kind of feature reuse, which to some extent enlarges network capacity without significantly increasing complexity.
%half of feature channels directly go through the entire residual layer and join into next unit. This can be regarded as a kind of feature reuse, which to some extent enlarges network capacity without significantly increasing complexity.

\begin{table}[!t]
\tabcolsep 2.1mm \caption{The architecture of LEDNet. ``Size'' denotes the dimension of output feature maps, $C$ is the number of classes.}
\begin{center}
\begin{tabular}{|c||l|c|}
\hline
\textbf{Stage} & \textbf{Type} &\textbf{Size} \\
\hline
\hline
\multirow{13}*{\rotatebox{90} {Encoder}}
&\textbf{Downsampling Unit}     &$512 \times 256 \times 32$ \\		
&$3 \times$ \textbf{SS-nbt Unit} &$512 \times 256 \times 32$ \\
&\textbf{Downsampling Unit}     &$256 \times 128 \times 64$ \\
&$2 \times$ \textbf{SS-nbt Unit} &$256 \times 128 \times 64$ \\
&\textbf{Downsampling Unit}     &$128 \times 64 \times 128$ \\
&\textbf{SS-nbt Unit} (dilated $r = 1$)  &$128 \times 64 \times 128$ \\
&\textbf{SS-nbt Unit} (dilated $r = 2$)  &$128 \times 64 \times 128$ \\
&\textbf{SS-nbt Unit} (dilated $r = 5$)  &$128 \times 64 \times 128$ \\
&\textbf{SS-nbt Unit} (dilated $r = 9$)  &$128 \times 64 \times 128$ \\
&\textbf{SS-nbt Unit} (dilated $r = 2$)  &$128 \times 64 \times 128$ \\
&\textbf{SS-nbt Unit} (dilated $r = 5$)  &$128 \times 64 \times 128$ \\
&\textbf{SS-nbt Unit} (dilated $r = 9$)  &$128 \times 64 \times 128$ \\
&\textbf{SS-nbt Unit} (dilated $r = 17$) &$128 \times 64 \times 128$ \\ \hline
\multirow{3}*{\rotatebox{90} {Decoder}}
&\multirow{2}{*}{\textbf{APN Module}}    &\multirow{2}{*}{$128 \times 64 \times C$} \\
&                               & \\
&\textbf{Upsampling Unit} ( $\times 8$) &$1024 \times 512 \times C$ \\
\hline
\end{tabular}
\end{center}\label{tab:LEDNet}
\end{table}

\subsection{LEDNet Architecture Designment}

%\noindent \textbf{LEDNet Architecture Designment.}
As shown in Table \ref{tab:LEDNet}, our LEDNet follows an encoder-decoder architecture. Unlike \cite{Badrinarayanan2015Segnet}, our approach employs an asymmetric sequential architecture, where a encoder produces downsampled feature maps, and a subsequent decoder adopts APN that upsamples the feature maps to match input resolution.

Besides SS-nbt unit, the encoder also includes downsampling unit, which is performed by stacking two parallel outputs of a single $3 \times 3$ convolution with stride 2 and a Max-pooling. Downsampling enables more deeper network to gather context, while at the same time helps to reduce computation. Note we postpone downsampling in encoder, in the similar spirit of \cite{Szegedy2016rethinking}. Moreover, the usage of dilated convolutions \cite{Romera2018erfnet,multi2016yu} allows our architecture to have large receptive field, leading to an improvement in accuracy. Compared to the use of larger kernel sizes, this technique has been proven more effective in terms of computational cost and parameters.

Inspired by attention mechanism \cite{hu2017squeeze}, our decoder designs a APN to perform dense estimation using spatial-wise attention. To increase receptive field, the APN adopts a pyramid attention module, which integrates features from three different pyramid scales. As shown in Figure \ref{fig:Overview}, we first utilize $3 \times 3$, $5 \times 5$, and $7 \times 7$ convolution with stride 2 to form multi-scale feature pyramid. Then the pyramid structure fuses information of different scales step-by-step, which can incorporate neighbor scales of context more precisely. Since high-level feature maps has small resolution, using large kernel size does not bring too much computation burden. Thereafter, a $1 \times 1$ convolution is applied to the output of encoder, then the convolutional feature maps are pixel-wisely multiplied by the pyramid attention features. To further enhance performance, a global average pooling branch is introduced to integrate global context prior attention. Finally, an upsampling unit is employed to match the resolution of input image. Benefiting from pyramid architecture, APN can capture multi-scale context cues, and produce pixel-level attention for convolutional features. Unlike DeepLab \cite{Chen2016deeplab} and PSPNet \cite{zhao2017pyramid} that stack multi-scale feature maps, our context information is pixel-wisely multiplied with original convolutional features, without introducing too much computational budgets.

\begin{table}[!t]
\tabcolsep 1.2mm \caption{Comparison with the state-of-the-art approaches in terms of segmentation accuracy and implementing efficiency.}
\begin{center}
\begin{tabular}{|c||ccccc|}
\hline
Method & Cla &Cat &Time(ms) &Speed(Fps) &Para(M) \\
\hline
\hline
SegNet\cite{Badrinarayanan2015Segnet}  &57.0 &79.1 &67 &15 &29.5\\
ENet\cite{Paszke2016enet}              &58.3 &80.4 &34 &31 &\textbf{0.36}\\
ESPNet\cite{Mehta2018espnet}           &60.3 &82.2 &\textbf{9} &\textbf{112} &0.40\\
CGNet\cite{wu2018cgnet}                &64.8 &85.7 &20 &50 &0.50\\
%ERFNet \cite{Romera2018erfnet}         &66.3 &85.2 &                             \\
ICNet \cite{Zhao2018ICnet}             &69.5 &86.4 &33 &30 &7.80\\
\hline
Ours                                   &\textbf{70.6} &\textbf{87.1} &14 &71 &0.94\\
\hline
\end{tabular}
\end{center}\label{tab:Result1}
\end{table}

\section{Experiments}\label{sec:experiments}

\subsection{Implementation Details}\label{sec:Implementation}

We select widely-used CityScapes dataset \cite{Cordts2016the} to evaluate our LEDNet, which includes 19 object categories and one additional background. Beside the images with fine pixel-level annotations that contain 2,975 training, 500 validation and 1,525 testing images, we also use the 20K coarsely annotated images for training. We adopt mean intersection-over-union (mIoU) averaged across all classes and categories to evaluate segmentation accuracy, while running time, speed (FPS), and model size (number of parameters) to measure implementing efficiency. To show the advantages of LEDNet, we selected 6 state-of-the-art lightweight networks as baselines, including SegNet \cite{Badrinarayanan2015Segnet}, ENet \cite{Paszke2016enet}, ERFNet \cite{Romera2018erfnet}, ICNet \cite{Zhao2018ICnet}, CGNet \cite{wu2018cgnet}, and ESPNet \cite{Mehta2018espnet}. For fair comparison, all the methods are conducted on the same hardware platform of Dell workstation with a single GTX 1080Ti GPU. We favor a large minibatch size (set as 5) to make full use of the GPU memory, where the initial learning rate is $5 \times 10^{-4}$ and the `poly' learning rate policy is adopted with power 0.9, together with momentum and weight decay are set to 0.9 and $10^{-4}$, respectively.

\begin{table*}[!t]
\tabcolsep 0.65mm \caption{Individual category results on the CityScapes test set in terms of class and category mIoU scores. Methods trained using both fine and coarse data are marked with superscript `${\dag}$'.}
\begin{center}
\begin{tabular}{|c||ccccccccccccccccccc||cc|}
\hline
Method  &{Roa}  &{Sid}  &{Bui}  &{Wal}  &{Fen}  &{Pol}  &{TLi}  &{TSi}  &{Veg}  &{Ter}  &{Sky}  &{Ped}  &{Rid}  &{Car}  &{Tru}  &{Bus}  &{Tra}  &{Mot}  &{Bic}  &{Cla}  &{Cat}\\
\hline
\hline
SegNet \cite{Badrinarayanan2015Segnet}  &96.4 &73.2 &84.0 &28.4 &29.0 &35.7 &39.8 &45.1 &87.0 &63.8 &91.8 &62.8 &42.8 &89.3 &38.1 &43.1 &44.1 &35.8 &51.9 &57.0 &79.1\\
ENet \cite{Paszke2016enet}              &96.3 &74.2 &75.0 &32.2 &33.2 &43.4 &34.1 &44.0 &88.6 &61.4 &90.6 &65.5 &38.4 &90.6 &36.9 &50.5 &48.1 &38.8 &55.4 &58.3 &80.4\\
ESPNet \cite{Mehta2018espnet}           &97.0 &77.5 &76.2 &35.0 &36.1 &45.0 &35.6 &46.3 &90.8 &63.2 &92.6 &67.0 &40.9 &92.3 &38.1 &52.5 &50.1 &41.8 &57.2 &60.3 &82.2\\
CGNet \cite{wu2018cgnet}                &95.5 &78.7 &88.1 &40.0 &43.0 &54.1 &59.8 &63.9 &89.6 &67.6 &92.9 &74.9 &54.9 &90.2 &44.1 &59.5 &25.2 &47.3 &60.2 &64.8 &85.7\\
ERFNet \cite{Romera2018erfnet}          &97.2 &\textbf{80.0} &89.5 &41.6 &45.3 &56.4 &60.5 &64.6 &91.4 &\textbf{68.7} &94.2 &76.1 &\textbf{56.4} &92.4 &45.7 &60.6 &27.0 &48.7 &61.8 &66.3 &85.2\\
ICNet \cite{Zhao2018ICnet}              &97.1 &79.2 &89.7 &43.2 &48.9 &61.5 &60.4 &63.4 &91.5 &68.3 &93.5 &74.6 &56.1 &\textbf{92.6} &51.3 &\textbf{72.7} &51.3 &\textbf{53.6} &70.5 &69.5 &86.4\\
\hline
Ours                                    &97.1 &78.6 &90.4 &46.5 &48.1 &60.9 &60.4 &71.1 &91.2 &60.0 &93.2 &74.3 &51.8 &92.3 &61.0 &72.4 &51.0 &43.3 &70.2 &69.2 &86.8\\
Ours$^{\dag}$                           &\textbf{98.1} &79.5 &\textbf{91.6} &\textbf{47.7} &\textbf{49.9} &\textbf{62.8} &\textbf{61.3} &\textbf{72.8} &\textbf{92.6} &61.2 &\textbf{94.9} &\textbf{76.2} &53.7 &90.9 &\textbf{64.4} &64.0 &\textbf{52.7} &44.4 &\textbf{71.6} &\textbf{70.6} &\textbf{87.1}\\
\hline
\end{tabular}
\end{center}\label{tab:Result2}
\end{table*}

\begin{figure*}[!t]
\centerline{\includegraphics[width = 1.0\textwidth]{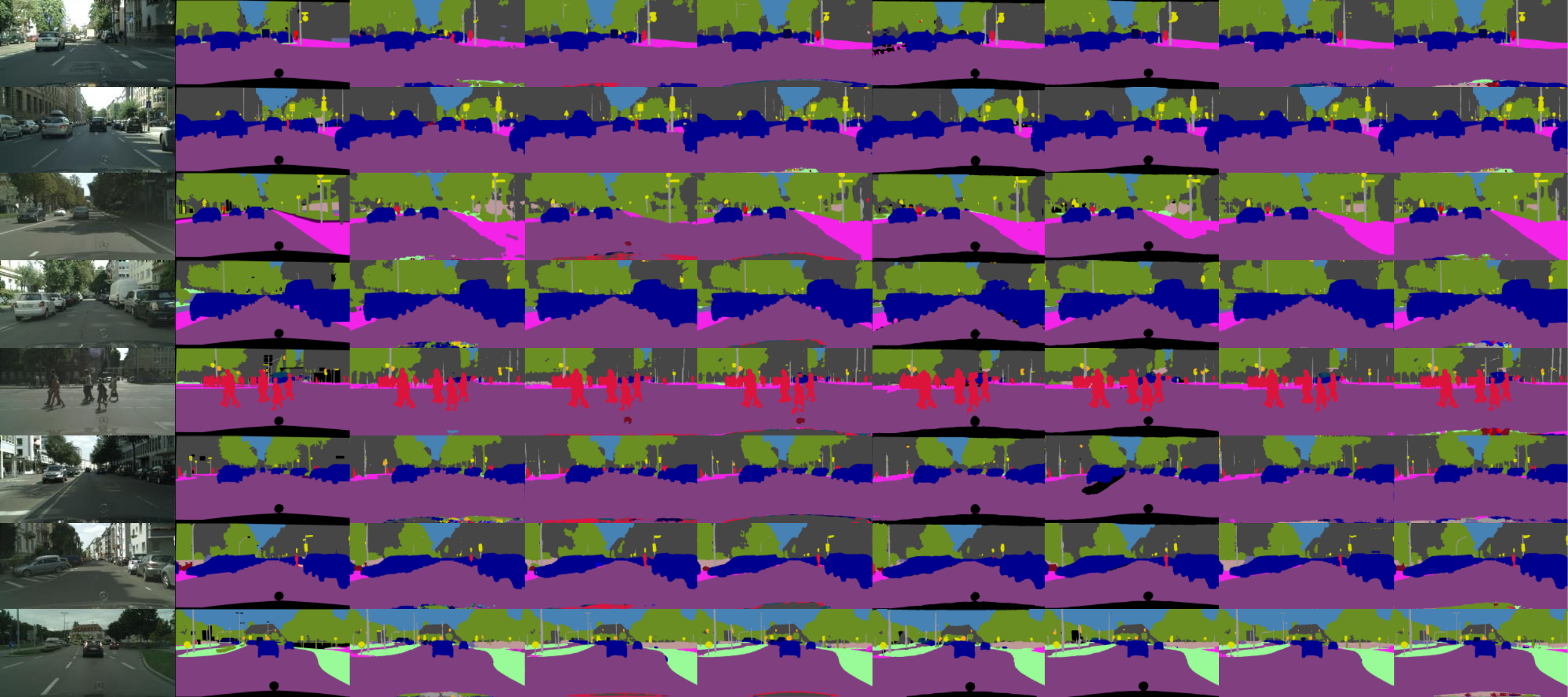}}
\caption{The visual comparison on CityScapes val dataset. From left to right are input images, ground truth, segmentation outputs from SegNet \cite{Badrinarayanan2015Segnet}, ENet \cite{Paszke2016enet}, ERFNet \cite{Romera2018erfnet}, ESPNet \cite{Mehta2018espnet}, ICNet \cite{Zhao2018ICnet}, CGNet \cite{wu2018cgnet}, and our LEDNet. (Best viewed in color)} \label{fig:Result3}
\end{figure*}

\subsection{Evaluation Results}\label{sec:Results}

Table \ref{tab:Result1} and Table \ref{tab:Result2} report comparison results, demonstrating that LEDNet achieves the best available trade-off in terms of accuracy and efficiency. Among all the approaches, our LEDNet yields 70.6\% class mIoU and 87.1\% category mIoU, respectively, where 13 out of the 19 categories obtains best scores. Regarding to the efficiency, LEDNet is nearly $5 \times$ faster and $30 \times$ smaller than SegNet \cite{Badrinarayanan2015Segnet}. Although ENet \cite{Paszke2016enet}, an another efficient network, is $1.5 \times$ efficient, and has  $3 \times$ less parameters, but delivers poor segmentation accuracy of 10\% drop than our LEDNet. Figure \ref{fig:Result3} shows some visual examples of segmentation outputs on the CityScapes validation set. It is demonstrated that, compared with baselines, our LEDNet not only correctly classifies object with different scales, but also produces consistent qualitative results for all classes.

\section{Conclusion and Future Work}\label{sec:conclusion}

This paper has described a LEDNet model, which designs an asymmetric encoder-decoder architecture for real-time semantic segmentation. The encoder adopts channel split and shuffle operations in residual layer, enhancing information communication in the manner of feature reuse. On the other hand, the decoder employs a APN, where the spatial pyramid structure is beneficial to enlarge receptive fields without introducing significant computational budgets. The entire network is trained end-to-end. The experimental results show our LEDNet achieves best trade-off on CityScapes dataset in terms of segmentation accuracy and implementing efficiency. The future work includes decomposing standard convolution in APN into 1D convolution, resulting in further lightweight network while still remaining segmentation accuracy.

%Through constructing contextual integration network, our LEDNet provides a more powerful representation that combines feature maps with different receptive fields. We evaluate our LEDNet on CityScapes dataset. The experimental results show the superior performance of our LEDNet in terms of efficiency and accuracy trade-off. The future work includes decomposing standard convolution in APN into point-wise convolution and depthwise separable convolution, resulting in further lightweight network while still remaining segmentation accuracy.

% References should be produced using the bibtex program from suitable
% BiBTeX files (here: strings, refs, manuals). The IEEEbib.bst bibliography
% style file from IEEE produces unsorted bibliography list.
% -------------------------------------------------------------------------
\bibliographystyle{IEEEbib}
\bibliography{strings,refs}

\end{document}